\documentclass[a4paper]{article}

\usepackage{INTERSPEECH2019}
\usepackage{url}

\title{Predicting the Leading Political Ideology of YouTube Channels\\ Using Acoustic, Textual, and Metadata Information}
\name{Yoan Dinkov$^1$, Ahmed Ali$^2$, Ivan Koychev$^1$, Preslav Nakov$^2$}
\address{
  $^1$Faculty of Mathematics and Informatics, Sofia University ``St Kliment Ohridski''\\
  $^2$Qatar Computing Research Institute, HBKU}
\email{\{jdinkov,koychev\}@uni-sofia.bg, \{amali,pnakov\}@hbku.edu.qa}

\begin{document}

\maketitle

\begin{abstract}

We address the problem of predicting the leading political ideology, i.e., left-center-right bias, for YouTube channels of news media. 
Previous work on the problem has focused exclusively on text and on analysis of the language used, topics discussed, sentiment, and the like. In contrast, here we study videos, which yields an interesting multimodal setup. Starting with gold annotations about the leading political ideology of major world news media from Media Bias/Fact Check, we searched on YouTube to find their corresponding channels, and we downloaded a recent sample of videos from each channel. We crawled more than 1,000 YouTube hours along with the corresponding subtitles and metadata, thus producing a new multimodal dataset.  We further developed a multimodal deep-learning architecture for the task.
Our analysis shows that the use of acoustic signal helped to improve bias detection by more than 6\% absolute over using text and metadata only. We release the dataset to the research community, hoping to help advance the field of multi-modal political bias detection.

\end{abstract}

\section{Introduction}
\label{sec:introduction}

Many of the issues discussed in the media today are deeply polarizing and thus are subject to political ideology or bias. On the one hand, it is natural for media that are openly associated with a particular political ideology to  offer strong cues about their political preferences when discussing such polarizing topics~\cite{kulkarni2018multi,P17-1068}. For example, what liberals and liberal media call the ``estate tax'', conservatives call the ``death tax''~\cite{P14-1105}. On the other hand, such left-vs-right (and other) biases can potentially exist in any news media, even in such that do not openly subscribe to a left/right agenda, and prefer to be seen as fair and balanced.
Although objectivity remains an important principle of journalism, researchers and watchdog groups agree that many of the supposedly objective news media are actually biased~\cite{niven2003objective,gentzkow2010drives}.

Spotting a systematic bias of a target news medium is easy for trained experts, and in many cases can be done by ordinary readers, but it requires exposure to a certain number of articles by the target medium. There have been systematic efforts to do this based on clear criteria by projects such as the Media Bias/Fact Check\footnote{\url{http://mediabiasfactcheck.com}} (MBFC) and AllSides\footnote{\url{http://www.allsides.com}}, to mention just two. However, as checking the bias is a tedious process, MBFC so far only covers 2,700 media, while this number is 600 for AllSides. Obviously, this does not scale well, and it is of limited utility if we wanted to characterize newly created media, so that readers are aware of what they are reading.

\noindent An attractive alternative is to try to automate the process, and there have been several attempts to do this in previous work. However, most work has focused on predicting the bias of individual articles~\cite{kulkarni2018multi,Horne:2018:ANL:3184558.3186987,DBLP:journals/corr/PotthastKRBS17} rather than characterizing entire news outlet~\cite{baly2018predicting,source:multitask:NAACL:2019}. Yet, we believe that focusing on entire news outlets is more useful if we want to raise awareness about what people are reading, and it is also arguably easier to work at the medium level as detecting political ideology requires looking for systematic bias over a period of time, for which looking at a single article is clearly not enough. 

Notably, previous work on bias detection has focused exclusively on newspaper-like media, which are text-based. However, online news are currently increasingly being consumed as multimedia, including videos. As a result, many media started creating YouTube channels where they are posting videos online. Thus, we believe that we should also move media bias analysis to YouTube channels, as for many media YouTube has become one of their main channels for content distribution. Moreover, videos allow us to explore not only the textual content of what is being said, but also the way it is said. Previous work has looked into text only, while we believe there is a lot that analysis of the acoustic signal can offer. Thus, below we aim to bridge this gap by combining textual and acoustic aspects of videos in order to predict the leading political ideology of YouTube channels. Our contributions are as follows:

\begin{itemize}
  \item We study an under-explored but arguably important problem: predicting the leading political ideology of YouTube channels.
  \item Unlike previous work, we model both the textual content and the acoustic signal (and metadata).
  \item As ours is a pioneering work, we create a new dataset of YouTube channels annotated for left-center-right bias, and we release the dataset and our code, which should facilitate future research.
\end{itemize}

\section{Related Work}
\label{sec:related_work}

In previous work, political ideology, also known as media bias, was used as a feature for ``fake news'' detection \cite{DBLP:journals/corr/abs-1803-10124}.
It has also been the target of classification, e.g., to predict whether an article is biased (\emph{political} or \emph{bias}) vs. unbiased \cite{Horne:2018:ANL:3184558.3186987}. 
Similarly, \cite{DBLP:journals/corr/PotthastKRBS17} classified the bias in a target article as
(\emph{i})~left vs. right vs. mainstream, or as
(\emph{ii})~hyper-partisan vs. mainstream. 
Left-vs-right bias classification at the article level was also explored by \cite{kulkarni2018multi}, who modeled both the textual and the URL contents of the target article. There has been also work characterizing the bias of an entire news outlet \cite{baly2018predicting,source:multitask:NAACL:2019}.
Bias has been also tragetted at the phrase and the sentence level \cite{P14-1105}, focusing on political speeches \cite{D13-1010}, legislative documents \cite{Gerrish:2011:PLR:3104482.3104544}, or targeting users in Twitter \cite{P17-1068}.

\noindent Another line of related work focuses on propaganda, which can be seen as a form of extreme bias \cite{rashkin-EtAl:2017:EMNLP2017,AAAI2019:proppy,Barron:19}.
See also a recent position paper \cite{Pitoura:2018:MBO:3186549.3186553} and an overview paper on bias on the Web \cite{Baeza-Yates:2018:BW:3229066.3209581}. 
Unlike the above work, here we focus on predicting the political ideology of YouTube channels. 
Moreover, most of the above work has analyzed text only, while we also use acoustics and meta data.

\section{Data}
\label{sec:dataset}

There is no pre-existing political labelling for videos or video channels on YouTube. Thus, following \cite{baly2018predicting,source:multitask:NAACL:2019}, we used media-level annotations for political bias from Media Bias/Fact Check (MBFC). MBFC uses the following seven categories: \emph{extreme left}, \emph{left}, \emph{center-left}, \emph{center}, \emph{center-right}, \emph{right}, and \emph{extreme right}. However, we found the \emph{center-left} and the \emph{center-right} labels confusing (are they more center or more left/right?), and therefore we dropped all instances with these labels. Moreover, in order to reduce the impact of subjective decisions made by the annotators, we merged the ``extreme'' examples with those that share the same polarity. Thus, ultimately, we model bias on a 3-point scale: \emph{left}, \emph{center}, and \emph{right}.
Table~\ref{tab:examples} shows some examples of media with their bias labels.

\begin{table}[tbh]
\small{
\centering
\begin{tabular}{@{ }@{ }l@{ }@{ }@{ }l@{ }@{ }@{ }l@{ }@{ }}
\toprule
\bf Name & \bf Bias & \multicolumn{1}{c}{\bf Youtube} \\ 
\midrule
Daily Mirror & \emph{left} & \url{~/user/dailymirror}\\
Associated Press & \emph{center} & \url{~/user/AssociatedPress} \\
Fox News & \emph{right} & \url{~/user/FoxNewsChannel}\\
\bottomrule
\end{tabular}}
\caption{Example media with their bias and Youtube channel.\label{tab:examples}}
\end{table}

We associated media with YouTube channels by looking for links to Youtube on the medium's home page. Unfortunately, the process could not be fully automated as many media had fully functioning YouTube channels without mentioning them in their websites and some media referred to external YouTube channels that were not theirs. Eventually, we had to process over 2,000 media manually by verifying the extracted links from the home page or by matching media to a channel after searching in YouTube. In particular, we tried to match the name of the medium (or its website) to the name of the YouTube channel, in addition to matching the logos, the font styles, the overall design, and the contact information. Overall, we managed to connect 1,161 media to their Youtube channel. We filtered 28 of them as they contained primarily non-English content. After additional filtering based on minimum duration for the videos and the captions, we ended up with a dataset of 421 channels. Table \ref{tab:dataset_information} shows some statistics about the dataset.

\begin{table}[htb]
    \centering
    \begin{tabular}{lr}
    \toprule
    \textbf{Name}  &\textbf{Value} \\
    \midrule
    Channels                      &  421 \\
    -- \emph{left} & 101 \\
    -- \emph{center} & 177 \\
    -- \emph{right} & 143 \\
    Videos                        &  3,345 \\
    Speech episodes               &  15,945 \\
    Average number of videos per channel     &  7.94 \\
    Average number of speech episodes per video   &  4.76 \\
    Average video duration              &  602 sec. \\
    \bottomrule
    \end{tabular}
    \caption{Statistics about our dataset.\label{tab:dataset_information}}
\end{table}

\noindent The dataset contains the following textual and metadata information about each video:
\begin{itemize}
    \item \emph{Text}: For each YouTube channel and for each Youtube video, we have a title as a mandatory text property. For channels, we can also have description. For videos, we have a description and tags. 
    \item \emph{Metadata}: There is no mandatory meta information, as it is generated based on Youtube statistics. For Youtube channels, we have a total number of views, a total number of videos, and subscribers count. For videos, there are five numerical values: number of views, number of likes, number of dislikes, number of comments, and duration (measured in seconds). In the experiments in Section 4, we use metadata for the videos only.
\end{itemize}

As we have an audio file and the captions for a video, we can match the location of the speech episodes on the audio timeline. Each caption has a starting and ending time and even though it does not match the actual speech (as it is used by YouTube captions API to signal when to display a caption and when to hide it in their video player), we found that about 80\% of these time intervals were filled with actual speech. Using this information, we retrieved the first five speech episodes for each audio file, requiring that the time between speech episodes be at least one second. Each speech episode is 15 seconds long. Overall, we generated 15,945 speech episodes from the 3,345 videos for our 421 Youtube channels.

\section{Model}

We use the following types of features:

\begin{itemize}
    \item NELA: We use features from the NELA toolkit \cite{horne2018assessing}, which were previously shown useful for detecting fake news, political bias, etc. The toolkit implements 130 features, which we extract separately from the title and from the description of the video: a total of 260 features.
    \item openSMILE: We experimented with the following predefined openSMILE configurations:
        \begin{itemize}        
            \item IS09\_emotion\cite{schuller2009interspeech}: The baseline from the INTERSPEECH'2009 Emotion Challenge.
            \item IS12\_speaker\_trait\cite{schuller2012interspeech}: The baseline from the INTERSPEECH'2012 Speaker Trait Challenge. This set of features yielded no actual improvement, and thus we do not use them in the experiments we report below (yet, we release them with the dataset).
        \end{itemize}
    \item
    i--vectors \cite{ali2015automatic}:
    These features model speech using a universal background model (UBM), which is typically a large Gaussian Mixture Model (GMM), trained on a large amount of data to represent general feature characteristics, which plays the role of a prior on how all speech styles look like. The i--vector approach is a powerful technique that summarizes all the updates happening during the adaptation of the UBM mean components to a given utterance. All this information is modeled in a low-dimensional subspace referred to as the total variability space. In the i--vector framework, each speech utterance can be represented by a GMM supervector. The i--vector is the low-dimensional representation of an audio recording that can be used for classification and estimation purposes. In our experiments, we used 600--dimensional i--vectors, which we trained using a GMM with 2048 components and BN features.
    
    \item BERT: In 2018 Google, presented a new model \cite{devlin2018bert} for sentence representation, which achieved very strong results on eleven natural language processing tasks including GLUE, MultiNLI, and SQuAD. Since then, it was used to improve over the state of the art for a number of NLP tasks. We used the cased pretrained model, and BERT-as-a-service \cite{xiao2018bertservice}, which generates a vector of 768 numerical values for a given text. 
    We generated features separately (\emph{i})~from the video's title, description and tags combined, and (\emph{ii})~from the video's captions.
\end{itemize}    

Table \ref{tab:features} gives some statistics about our feature set.

\begin{table}[th]
    \centering
    \begin{tabular}{lr}
    \toprule
    \textbf{Type}  &\textbf{Features} \\
    \midrule
    BERT (captions)                     & 768 \\
    BERT (title, description, tags)     &  768 \\
    i--vectors (Speech embeddings)        & 600 \\
    NELA (title, description)           & 260 \\ 
    Numeric (metadata)                  & 5 \\
    openSMILE (IS09 emotion)           & 385 \\
    \bottomrule
    \end{tabular}
    \caption {\label{tab:features}Statistics about our feature set.}
\end{table}

\section{Experiments and Evaluation} 

\subsection{Experimental Setup}

We used stratified 5-fold cross-validation at the YouTube channels level. We further split the channels into videos, and the videos into episodes. Then, we extracted features from each episode, we aggregated these features at the video level, and we performed classification using distant supervision, i.e., assigning to each video the label of the channel it comes from.
Finally, we aggregated the posterior probabilities to obtain a probability distribution over the bias labels but now for channels.

\begin{figure}[tbh]
  \centering
  \includegraphics[width=8cm]{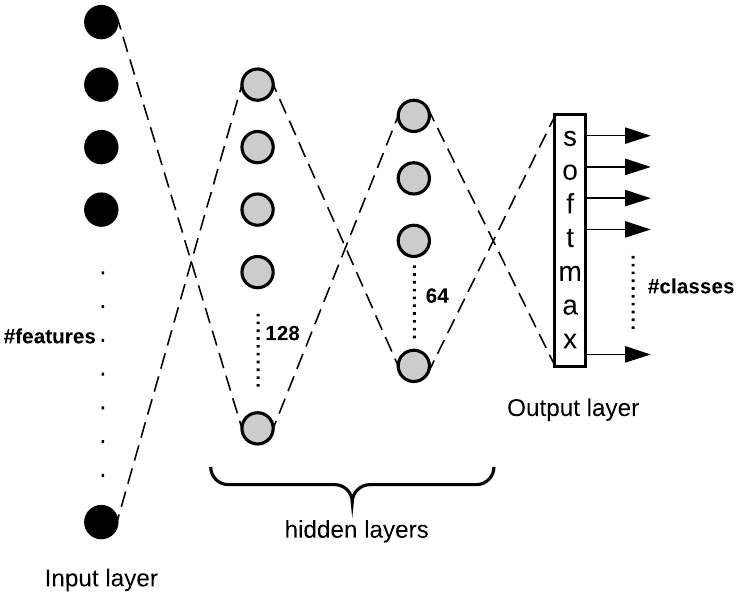}
  \caption{\label{fig:nn_diagram}The neural model architecture.}
\end{figure}

For classification, we used a feed-forward neural network with two hidden layers (128 nodes with ReLU activation, and then 64 nodes with tanh activation), and dropout layers (with 0.2 dropout rate) before each layer, as shown in Figure~\ref{fig:nn_diagram}. For optimization, we used Adagrad with a batch size of 75, and we ran it for 35 epochs.

\noindent We first experimented with each feature type in isolation. Then, we tried various combinations thereof. The results are shown in Table~\ref{tab:results}.

\begin{itemize}
    \item \emph{Baseline}: This is a majority class baseline, where we predict the most common label in the dataset, which is \emph{center} (see Table~\ref{tab:dataset_information}). This baseline yields 42\% accuracy.
    \item \emph{Individual feature types}: We perform isolated experiments with each of the feature types defined in Section~\ref{sec:dataset} (and also listed in Table~\ref{tab:features}). We can see that the best-performing features are those based on BERT, achieving up to 68.91\% accuracy for BERT that is calculated on title + description + tags. BERT based on captions comes second with accuracy of 64.64\%. The two types of audio features, based on i--vectors and on openSMILE, perform much worse with accuracy of 50.85\% and 56.63\%, respectively. The metadata features yield comparable accuracy of 50.35\%. Interestingly, the NELA features did not help at all, and they performed the same as the majority class baseline at 42.04\% accuracy;\footnote{The NELA features performed even worse when extracted from the captions, going below the baseline. Thus we did not include this result in Table~\ref{tab:results}.} this is a surprising result that is worth a closer investigation in future work.
    \item \emph{Feature combinations}: We performed various combinations of feature types, and we report some of the most interesting ones below. We can see that combining the textual and the metadata features yields 70.10\% accuracy, which is about 1.2\% of improvement absolute over the best text-only feature. Adding the i--vectors to the combination gives a marginal improvement to 70.33\%. Adding both audio features yields an improvement to 72.02\%. However, using only openSMILE and not using the i--vectors yielded the best result: 73.42\% accuracy. The fact that the i--vectors did not help is worth a closer investigation in future work. Comparing line 11 to line 7, we can see that the feature combinations yield 4.5\% improvement absolute. Comparing line 10 to line 8, we can conclude that using audio information yields about 2\% of absolute improvement, which shows the importance of modeling the acoustics as an additional information source, even if the audio features are relatively weak in isolation (see lines 4 and 5).
\end{itemize}

\begin{table}[th]
  \centering
  \begin{tabular}{@{ }@{ }l@{ }@{ }@{ }@{ }l@{ }@{ }@{ }@{ }l@{ }@{ }@{ }c@{ }@{ }}
    \toprule
    \textbf{\#} & \textbf{Type} &\textbf{Experiment} &\textbf{Accuracy} \\
    \midrule
    1 &                 &  Baseline                                 & 42.04 \\
    \hline
    2 & Text            &  NELA (title, description)                &   42.04 \\
    3 & Meta            &  Numerical                                &   50.35 \\
    4 & Audio           &  i--vectors                        &   50.85 \\
    5 & Audio           &  openSMILE                                &   56.63 \\
    6 & Text            &  BERT (captions)                          &   64.64 \\
    7 & Text            &  BERT (title, description, tags)          &   68.91 \\
    \hline
    8 & Combined        &  Text + Meta                              &   70.10 \\
    9 & Combined        &  Text + Meta + i--vectors         &   70.33 \\
    10 & Combined        &  Text + Meta + Audio                      &   72.02 \\
    \textbf{11} & \textbf{Combined} &\textbf{Text + Meta + openSMILE}        &\textbf{73.42} \\
    \bottomrule
  \end{tabular}
  \caption{\label{tab:results}Evaluation results.}
\end{table}

\section{Discussion}
\label{sec:discussion}
 
\subsection{Aggregation Strategies}

In the above experiments, we were splitting the channels into videos, and then the videos into episodes. Then, we were extracting features from the episodes, which we were averaging to form feature vectors for the videos. Next, we were training a classifier and we were making predictions at the video level using distant supervision, i.e.,~assuming each video has the same bias as the Youtube channel it came from. Finally, we were aggregating, i.e.,~averaging, the posterior probabilities for the videos from the same channel to make a prediction for the bias of that channel.
Two natural questions arise about this setup: (\emph{i})~Why not perform the classification at the episode level and then aggregate the posteriors from the classification for \emph{episodes} rather than for \emph{videos}? (\emph{ii})~Why not use a different aggregation strategy to perform the aggregation of the predictions, e.g.,~why not try \emph{maximum} instead of \emph{average}?

To answer these questions, we performed an ablation study for all our experiments trying classification at the \emph{episode} vs. the \emph{video} level, and aggregation using \emph{maximum} vs. \emph{average}.
A potential advantage of using \emph{maximum} is that it could be more sensitive to stromg signal coming from a single episode; however, it could also turn out to be too sensitive to a single episode. A possible advantage of classifying at the episode level is that there would be more training data; however, episodes are short, and thus it is harder to make clasification at that level.
Eventually, we found that the original strategy of classifying at the \emph{video} level and aggregating using \emph{average} performed best.

Table~\ref{tab:different_approaches} illustrates this as an ablation for our best result from Table~\ref{tab:results}. We can see from Table~\ref{tab:different_approaches} that using \emph{maximum} for aggregating the posteriors performs worse than using \emph{average}. We further see that classifying at the video level is better than classifying at the episode level (if we use average for aggregation). We can further see that our choice was better than the other three alternatives.

\begin{table}[th]
    \centering
    \begin{tabular}{llll}
    \toprule
    \textbf{\#}  &\textbf{Level}  &\textbf{Aggregation} &\textbf{Accuracy} \\
    \midrule
    \textbf{1} & \textbf{Video}  & \textbf{Average}&   \textbf{73.42} \\
    2 & Video           &  Maximum            &   67.94 \\
    3 & Episodes        &  Average            &   72.02 \\
    4 & Episodes        &  Maximum            &   71.27 \\
    \bottomrule
    \end{tabular}
    \caption {\label{tab:different_approaches}Results for different basic classification levels and for different kinds of aggregation.}
\end{table}

\subsection {Impact of i--vectors and openSMILE Features}

Next, we looked into why openSMILE worked better than i--vesctors. In Table \ref{tab:features}, we can see that there are 600 i-vector features and 385 features generated from openSMILE's IS09\_emotion configuration (explained in section \ref{sec:dataset}). Thus, one possible explanation is that i--vectors simply have more features, and we do not have enough training data to make use of so many features. 

The difference could be also due to openSMILE focusing on representing the emotions in a target speech episode, while i--vectors retrieve general feature characteristics from a target episode, and thus should be expected to be of limited utility for our task.

\section{Conclusion and Future Work}
\label{sec:future_work}

We have addressed the problem of predicting the leading political ideology, i.e., left-center-right bias, for YouTube channels of news media. 
Previous work on the problem has focused exclusively on by printed and online text media, and on analysis of the language used, topics discussed, sentiment, and the like. In contrast, here we studied videos, which yielded an interesting multimodal setup, where we have textual, acoustic, and metadata information (and also video, which can be analyzed in future work). We crawled more than 1,000 YouTube hours along with the corresponding subtitles and metadata, thus producing a new multimodal dataset.  We further developed a multimodal deep-learning neural network architecture for the task. This model achieved very sizable improvements over the baseline: accuracy of 73.42\% (baseline: 42.04\%).
Our analysis has shown that the use of acoustic signal helped to improve bias detection by more than 6\% absolute over using text and metadata only. We release the dataset to the research community, hoping to help advance the field of multi-modal political bias detection.

In future work, we plan to increase the size of the dataset in terms of processed volume. Currently, we have around 3,300 videos, and we plan to expand the data by several thousand more videos, aiming at 3,000-5,000 videos per class. We further plan to move to 5-way or 7-way classification by adding the \emph{center-left} and the \emph{center-right} labels, and possibly also \emph{extreme-left} and \emph{extreme-right}. Another direction for future work is better modeling of the audio features: we plan further experiments with i--vectors, trying various openSMILE configurations, and trying custom neural network architectures to extract better task-specific acoustic features. We further want to add features from the video itself, features from image analysis of the thumbnails, as well as social information from the comments in the Youtube forum below the target video. Last but not least, we want to develop end-to-end training in a single neural network.

\section{Contribution}
\label{sec:contribution}
The dataset we created can be found on Kaggle\footnote{\url{http://www.kaggle.com/yoandinkov/youtubepoliticalbias/}}.
We further open-sourced our code for the experiments described in this paper, and it can be found on GitHub\footnote{\url{http://github.com/yoandinkov/interspeech-2019}}. The GitHub repository contains additional information about our experimental setup and the packages used.

\section{Acknowledgements}

This research is part of the Tanbih project,\footnote{\url{http://tanbih.qcri.org/}} which aims to limit the effect of ``fake news'', propaganda and media bias by making users aware of what they are reading. The project is developed in collaboration between the Qatar Computing Research Institute (QCRI), HBKU and the MIT Computer Science and Artificial Intelligence Laboratory (CSAIL).

This research is also partially supported by Project UNITe BG05M2OP001-1.001-0004 funded by the OP ``Science and Education for Smart Growth'', co-funded by the EU through the ESI Funds.

\bibliographystyle{IEEEtran}
\bibliography{references}

\end{document}